\documentclass[11pt,a4paper]{article}
\usepackage[hyperref]{naaclhlt2019}
\usepackage{times}
\usepackage{latexsym}

\usepackage{caption}
\usepackage{amsmath}
\usepackage{amssymb}
\usepackage{amsfonts}

\usepackage{multirow}
\usepackage{tabularx}

\usepackage{boxedminipage} 

\usepackage{url}
\usepackage{verbatim}

\usepackage{booktabs}

\usepackage{xcolor,colortbl}

\usepackage{xspace}
\usepackage{graphicx}
\usepackage{subcaption}
\usepackage{arydshln}

\aclfinalcopy 

\newcommand{\red}[1]{\textcolor{red}{#1}}

\newcommand{\basespk}[0]{$S_0$\xspace}
\newcommand{\pragd}[0]{$S_1^D$\xspace}
\newcommand{\pragr}[0]{$S_1^R$\xspace}
\newcommand{\distr}[0]{\ensuremath\widetilde\imath}

\newcommand{\bleu}{\textsc{bleu}\xspace}
\newcommand{\nist}{\textsc{nist}\xspace}
\newcommand{\meteor}{\textsc{meteor}\xspace}
\newcommand{\cider}{\textsc{cide}r\xspace}
\newcommand{\rouge}{\textsc{rouge}\xspace}
\newcommand{\rougel}{\textsc{rouge-l}\xspace}

\newcommand{\phan}{\phantom{$^\dagger$}}

\newcommand{\ie}{i.e., }
\newcommand{\eg}{e.g., }

\definecolor{snsred}{HTML}{C44E52}
\definecolor{snsgreen}{HTML}{55a868}

\def\rot#1{\rotatebox{90}{#1}}

\title{Pragmatically Informative Text Generation}

\author{Sheng Shen$^\dagger$ ~ Daniel Fried$^\dagger$ ~ Jacob Andreas$^\ddagger$ ~ Dan Klein$^\dagger$ \\
$^\dagger$Computer Science Division, UC Berkeley\\ 
$^\ddagger$Computer Science and Artificial Intelligence Laboratory, MIT \\
{\tt \{sheng.s,dfried,klein\}@berkeley.edu, jda@mit.edu}  \\
}

\date{}

\begin{document}

\allowdisplaybreaks

\newcolumntype{R}{>{\raggedleft\arraybackslash}X}

\maketitle
\begin{abstract}
We improve the informativeness of models for conditional text generation using techniques from computational pragmatics. These techniques formulate language production as a game between speakers and listeners, in which a speaker should generate output text that a listener can use to correctly identify the original input that the text describes. While such approaches are widely used in cognitive science and grounded language learning, they have received less attention for more standard language generation tasks.  We consider two pragmatic modeling methods for text generation: one where pragmatics is imposed by information preservation, and another where pragmatics is imposed by explicit modeling of distractors. We find that these methods improve the performance of strong existing systems for abstractive summarization and generation from structured meaning representations.
\end{abstract}

\section{Introduction}

Computational approaches to pragmatics cast
language generation and interpretation as game-theoretic or
Bayesian
inference procedures \cite{Golland10Game,frank2012predicting}.
While such approaches are capable of modeling a variety of pragmatic phenomena,
their main application in natural language processing has been to
improve the \emph{informativeness} of generated text in grounded language learning problems \cite{monroe2018bilingual}.
In this paper, we show that pragmatic reasoning can be similarly used to improve performance
in more traditional language generation tasks like generation
from structured meaning representations (\autoref{fig:example_figure})
and summarization.

Our work builds on a line of learned Rational Speech Acts (RSA) models
\cite{Monroe15RationalSpeech,Andreas16Pragmatics}, in which generated strings
are selected to optimize the behavior of an embedded listener model. The
canonical presentation of the RSA framework \cite{frank2012predicting} is
grounded in reference resolution: models of speakers attempt to describe
referents in the presence of distractors, and models of listeners attempt to
resolve descriptors to referents. Recent work has extended these models to more complex groundings, including images
\cite{Mao15Generation} and trajectories
\cite{fried2017unified}. The techniques used in these settings are similar, and
the primary intuition
of the RSA framework is preserved: from the speaker's perspective, a good
description is one that picks out, as discriminatively as possible, the content
the speaker intends for the listener to identify.


\begin{figure}[t]
\begin{boxedminipage}{\columnwidth}
\scriptsize
\textbf{Input meaning representation ($i$):}
\newline
\textsc{Name[Fitzbillies], EatType[coffee shop],  Food[English], PriceRange[cheap], CustomerRating[5~out~of~5], Area[Riverside], FamilyFriendly[yes]}
\newline \rule{\columnwidth}{1.2pt}
\textbf{Human-written}
\newline
\emph{A cheap coffee shop in riverside with a 5 out of 5 customer rating is Fitzbillies. Fitzbillies is family friendly and serves English food.}
\newline \rule{\columnwidth}{0.4pt}
\textbf{Base sequence-to-sequence model ($S_0$)}
\newline
\emph{Fitzbillies is a family friendly coffee shop located near the river.}
\newline \rule{\columnwidth}{0.4pt}
\textbf{Distractor-based pragmatic system ($S^D_1$)}
\newline
\emph{Fitzbillies is a family friendly coffee shop that serves English food. It is located in riverside area. It has a customer rating of 5 out of 5 and is cheap.}
\newline \rule{\columnwidth}{0.4pt}
\textbf{Reconstructor-based pragmatic system ($S^R_1$)}
\newline
\emph{Fitzbillies is a family friendly coffee shop that serves cheap English food in the riverside area. It has a customer rating of 5 out of 5.}
\end{boxedminipage}
\caption{
\label{fig:example_figure}
Example outputs of our systems on the E2E generation task. While a base sequence-to-sequence model (\basespk, Sec.\ \ref{sec:tasks}) fails to describe all attributes in the input meaning representation, both of our pragmatic systems (\pragr, Sec.\ \ref{sec:pragr} and \pragd, Sec.\ \ref{sec:pragd}) and the human-written reference do.
\vspace{-1em}
}
\end{figure}


Outside of grounding, cognitive modeling \cite{Frank09PragmaticExperiments}, and targeted analysis of linguistic phenomena \cite{orita2015discourse}, rational speech acts models have seen limited application in the natural language processing
literature. In this work we show that they can be extended to a distinct class
of language generation problems that use as referents structured descriptions 
of lingustic content, or other natural language texts.
In accordance with the maxim of quantity \cite{Grice70Conversation} or the Q-principle \cite{Horn84Taxonomy},
pragmatic approaches naturally correct underinformativeness problems 
observed in state-of-the-art language generation systems ($S_0$ in \autoref{fig:example_figure}).

We present experiments on two language generation tasks: 
generation from meaning representations \cite{novikova2017e2e} and summarization.
For each task, we evaluate two models of pragmatics:
the reconstructor-based model of \citet{fried2017unified} and the distractor-based 
model of \citet{cohn2018pragmatically}. Both models improve performance on both tasks, increasing \rouge scores by 0.2--0.5 points on the CNN/Daily Mail  
abstractive summarization dataset and \bleu scores by 2 points on the 
End-to-End (E2E) generation dataset, obtaining new state-of-the-art results.

\section{Tasks}
\label{sec:tasks}

We formulate a conditional generation task as taking an input $i$ from a space of possible inputs $\mathcal{I}$
(\eg input sentences for abstractive summarization; meaning representations for structured generation) and producing an output $o$ as a sequence of tokens $(o_1, \ldots, o_T)$. 
We build our pragmatic approaches on top of learned \emph{base speaker} models \basespk, which produce a probability distribution $S_0(o \mid i)$ 
over output text for a given input.
We focus on two conditional generation tasks where the information in the input context should largely be preserved in the output text, and apply the pragmatic procedures outlined in Sec.\ \ref{sec:pragmatic_models} to each task. 
For these \basespk models we use systems from past work that are strong, but may still be underinformative relative to human reference outputs (\eg \autoref{fig:example_figure}).



\paragraph{Meaning Representations}
Our first task is generation from structured meaning representations (MRs) containing attribute-value pairs \cite{novikova2017e2e}. An example is shown in \autoref{fig:example_figure}, where systems must generate a description of the restaurant with the specified attributes. We apply pragmatics to encourage output strings from which the input MR can be identified. For our \basespk model, we use a publicly-released neural generation system \citep{puzikov2018e2e} that achieves comparable performance to the best published results in \citet{duvsek2018findings}.

\paragraph{Abstractive Summarization}
Our second task is multi-sentence document summarization. There is a vast amount of past work on summarization \cite{nenkova2011automatic}; recent neural models have used large datasets (\eg \citet{hermann2015teaching}) to train models in both the extractive \cite{cheng2016summarization,nallapati2017summarunner} and abstractive  \cite{rush2015abstractive,see2017summarization} settings.
Among these works, we build on the recent 
abstractive neural summarization system of \newcite{chen2018fast}. First, this system uses a sentence-level extractive model \textsc{rnn-ext} to identify a sequence of salient sentences $i^{(1)}, \ldots i^{(P)}$ in each source document. Second, the system uses an abstractive model \textsc{abs} to rewrite each $i^{(p)}$ into an output $o^{(p)}$, which are then concatenated to produce the final summary. We rely on the fixed \textsc{rnn-ext} model to extract sentences as inputs in our pragmatic procedure, using \textsc{abs} as our \basespk model and applying pragmatics to the $i^{(p)} \rightarrow o^{(p)}$ abstractive step.

\section{Pragmatic Models}
\label{sec:pragmatic_models}

To produce informative outputs, we consider pragmatic methods that extend the \emph{base speaker} models, \basespk, using \emph{listener} models, $L$, which produce a distribution $L(i \mid o)$ over possible inputs given an output. Listener models are used to derive \emph{pragmatic speakers}, $S_1(o \mid i)$, which produce output that has a high probability of making a listener model $L$ identify the correct input. 
There are a large space of possible choices for designing $L$ and deriving $S_1$; we follow two lines of past work which we categorize as \emph{reconstructor-based} and \emph{distractor-based}.
We tailor each of these pragmatic methods to both our two tasks by developing reconstructor models and methods of choosing distractors.

\subsection{Reconstructor-Based Pragmatics}
\label{sec:pragr}
Pragmatic approaches in this category \cite{duvsek2016sequence,fried2017unified} rely on a \emph{reconstructor listener} model $L^R$ defined independently of the speaker. This listener model produces a distribution $L^R(i \mid o)$ over 
all possible input contexts $i \in \mathcal{I}$, given an output description $o$. 
We use sequence-to-sequence or structured classification models for $L^R$ (described below), and train these models on the same data used to supervise the $S_0$ models. 

The listener model and the base speaker model together define a \emph{pragmatic speaker}, with output score given by:
\begin{equation}
\label{eq:reconstructor_s1}
S_1^R(o \mid i) = L^R(i \mid o)^{\lambda} \cdot S_0(o \mid i)^{1 - \lambda}
\end{equation}
where $\lambda$ is a \emph{rationality parameter} that controls how much the model optimizes for discriminative outputs (see \citet{monroe2017colors} and \citet{fried2017unified} for a discussion). 
We select an output text sequence $o$ for a given input $i$ by choosing the highest scoring output under Eq.\ \ref{eq:reconstructor_s1} from a set of candidates obtained by beam search in $S_0(\cdot \mid i)$.

\paragraph{Meaning Representations}
We construct $L^R$ for the meaning representation generation task as a multi-task, multi-class classifier, defining a distribution over possible values for each attribute.
Each MR attribute has its own prediction layer and attention-based aggregation layer, which conditions on
a basic encoding of $o$ shared across all attributes.
See Appendix \ref{sec:appendix_reconstructor} for architecture details.
We then define $L^R(i\mid o)$ as the joint probability of predicting all input MR attributes in $i$ from $o$.

\paragraph{Summarization}
To construct $L^R$ for summarization, we train an \textsc{abs} model (of the type we use for \basespk, \citet{chen2018fast}) but in reverse, \ie taking as input a sentence in the summary and producing a sentence in the source document. We train $L^R$ on the same heuristically-extracted and aligned source document sentences used to train \basespk \cite{chen2018fast}. 


\subsection{Distractor-Based Pragmatics}
\label{sec:pragd}
Pragmatic approaches in this category \cite{frank2012predicting,Andreas16Pragmatics,vedantam2017captions,cohn2018pragmatically} derive pragmatic behavior by producing outputs that distinguish the input $i$ from an alternate \emph{distractor} input (or inputs). 
We construct a distractor $\distr$ for a given input $i$ in a task-dependent way.\footnote{In tasks such as contrastive captioning or referring expression generation, these distractors are given; for the conditional generation task, we will show that pragmatic behavior can be obtained 
by constructing or selecting a single distractor that contrasts with the input $i$.}

We follow the approach of \newcite{cohn2018pragmatically}, outlined briefly here.
The base speakers we build on produce outputs incrementally, where the probability of $o_t$, the word output at time $t$, is conditioned on the input and the previously generated words: $S_0(o_t \mid i, o_{<t})$.
Since the output is generated incrementally and there is no separate listener model that needs to condition on entire output decisions, the distractor-based approach is able to 
make pragmatic decisions at each word rather than choosing between entire output candidates (as in the reconstructor approaches).

The listener $L^D$ and pragmatic speaker $S_1^D$ are derived from the base speaker $S_0$ and a belief  distribution $p_t(\cdot)$ maintained at each timestep $t$ over the possible inputs $\mathcal{I}^D$:
{
\begin{align}
\label{eq:distractor_l0}
L^D(i \mid o_{<t}) &\propto S_0(o_{<t} \mid i) \cdot p_{t-1}(i) \\
\label{eq:distractor_s1}
\hspace{-13pt} S^D_1(o_t \mid i, o_{<t}) &\propto L^D(i \mid o_{<t})^\alpha \cdot S_0(o_t \mid i, o_{<t}) \\
\label{eq:distractor_l1}
p_{t}(i) &\propto S_0(o_t \mid i, o_{<t}) \cdot L^D(i \mid o_{<t})
\end{align}
}%
where $\alpha$ is again a rationality parameter, and the initial belief distribution $p_0(\cdot)$ is uniform, \ie $p_0(i) = p_0(~\distr~) = 0.5$. 
Eqs.\ \ref{eq:distractor_l0} and \ref{eq:distractor_l1} are normalized over the true input $i$ and distractor $\distr$; 
Eq.\ \ref{eq:distractor_s1} is normalized over the output vocabulary. 
We construct an output text sequence for the pragmatic speaker $S^D_1$ incrementally using beam search to approximately maximize Eq.\ \ref{eq:distractor_s1}.

\paragraph{Meaning Representations}
A distractor MR is automatically constructed for each input 
to be the most distinctive possible against the input. We construct this distractor by masking each present input attribute and replacing the value of each non-present attribute with the value that is most frequent for that attribute in the training data. For example, for the input MR in \autoref{fig:example_figure}, the distractor is \textsc{Near[Burger King]}. 

\paragraph{Summarization}
For each extracted input sentence $i^{(p)}$, we use the previous extracted sentence $i^{(p-1)}$ 
from the same document as the distractor input $\distr$ (for the first sentence we do not use a distractor).  
This is intended to encourage outputs $o^{(p)}$ to contain distinctive information 
against other summaries produced 
within the same document.

\section{Experiments}

For each of our two conditional generation tasks we evaluate on a standard benchmark dataset, following past work by using automatic evaluation against human-produced reference text. We choose hyperparameters for our models (beam size, and parameters $\alpha$ and $\lambda$) to maximize task metrics on each dataset's development set; see Appendix~\ref{sec:appendix_hyperparameters} for the settings used.\footnote{Our code is 
publicly available at 
\url{https://github.com/sIncerass/prag\_generation}.
}

\subsection{Meaning Representations}
\begin{table}[t]
\small
\centering
 \hskip-2.4mm
\scalebox{0.96}{
\begin{tabular}{p{13mm}ccccc}
\toprule
  System & \bleu & \nist & \meteor & \textsc{r-l} & \cider \\ 
  \midrule
T-Gen & 65.93\phan & 8.61\phan & 44.83\phan & 68.50\phan & 2.23\phan  \\ 
Best Prev.\hspace{-4mm} & 66.19$^\dagger$ & 8.61$^\dagger$ & {\bf 45.29}$^\ddagger$ & {\bf 70.83}$^\diamond$ & 2.27$^\bullet$  \\ 
\midrule
\basespk & 66.52\phan & 8.55\phan & 44.45\phan & 69.34\phan & 2.23\phan   \\ 
\basespk $\times 2$ & 65.93\phan & 8.31\phan & 43.52\phan & 69.58\phan & 2.12\phan  \\ 
\pragr & {\bf 68.60}\phan & {\bf 8.73}\phan & {\bf 45.25}\phan & {\bf 70.82}\phan & {\bf 2.37}\phan  \\ 
\pragd & 67.76\phan & 8.72\phan & 44.59\phan & 69.41\phan & 2.27\phan  \\ 
\bottomrule
\end{tabular}
}
\caption{\label{tbl:e2e} Test results for the E2E generation task, in comparison to the T-Gen baseline \cite{duvsek2016sequence} and the best results from the E2E challenge, reported by \citet{duvsek2018findings}: $^\dagger$\citet{juraska2018ensemble}, $^\ddagger$\citet{puzikov2018e2e}, $^\diamond$\citet{zhang2018e2e}, and $^\bullet$\citet{gong2018e2e}. We bold our highest performing model on each metric, as well as previous work if it outperforms all of our models.
\vspace{-1em}
}
\end{table}

We evaluate on the E2E task of generation from meaning representations containing restaurant attributes \cite{novikova2017e2e}.
We report the task's five automatic metrics:  \bleu \cite{papineni2002bleu}, \nist \cite{doddington2002automatic}, \meteor \cite{lavie2007meteor}, \rougel \cite{lin2004rouge} and \cider \cite{vedantam2015cider}. 

\autoref{tbl:e2e} compares the performance of our base \basespk and pragmatic models to the baseline T-Gen system \cite{duvsek2016sequence} and the \emph{best previous} result from the 20 primary systems evaluated in the E2E challenge \cite{duvsek2018findings}. 
The systems obtaining these results encompass a range of approaches: a template system \cite{puzikov2018e2e}, a neural model \cite{zhang2018e2e}, models trained with reinforcement learning \cite{gong2018e2e}, and systems using ensembling and reranking \cite{juraska2018ensemble}.
To ensure that the benefit of the reconstructor-based pragmatic approach, which uses two models, is not due solely to a model combination effect, we also compare to an ensemble of two base models (\basespk $\times 2$). This ensemble uses a weighted combination of scores of two independently-trained \basespk models, following Eq.\ \ref{eq:reconstructor_s1} (with weights tuned on the development data).

Both of our pragmatic systems improve over the strong baseline \basespk system 
on all five metrics, with the largest improvements (2.1 \bleu, 
0.2 \nist,
0.8 \meteor, 1.5 \rougel, and 
0.1 \cider)
from the \pragr model. This \pragr model outperforms the previous best results obtained by any system in the E2E challenge on \bleu, \nist, and \cider, with comparable performance on \meteor and \rougel.

\subsection{Abstractive Summarization}

We evaluate on the CNN/Daily Mail summarization dataset \cite{hermann2015teaching,nallapati2016abstractive}, using See et al.'s~(\citeyear{see2017summarization}) non-anonymized preprocessing. As in previous work \cite{chen2018fast},  we evaluate using \rouge and \meteor.

\autoref{tbl:summarization} compares our pragmatic systems to the base \basespk model (with scores taken from \citet{chen2018fast}; we obtained comparable performance in our reproduction\footnote{We use retrained versions of \citet{chen2018fast}'s sentence extractor and abstractive \basespk models in all our experiments, as well as their n-gram reranking-based inference procedure, replacing scores from the base model \basespk with scores from \pragr or \pragd in the respective pragmatic procedures.}), an ensemble of two of these base models, and the \emph{best previous} abstractive summarization result for each metric on this dataset \cite{celikyilmaz2018communicating,paulus2018summarization,chen2018fast}.
We also report two extractive baselines: \emph{Lead-3}, which uses the first three sentences of the document as the summary \cite{see2017summarization}, and \emph{Inputs}, the concatenation of the extracted sentences 
used as inputs to our models (\ie $i^{(1)}, \ldots, i^{(P)}$).

\begin{table}[t]
\centering
\small
\begin{tabular}{lcccc}
\toprule
System & R-1 & R-2 & R-L & \meteor \\
\midrule
Extractive \\
\midrule
Lead-3 & 40.34\phan & 17.70\phan & 36.57\phan & {\bf 22.21}\phan \\
Inputs 
& 38.93\phan & 18.23\phan & 35.90\phan & {\bf 24.66}\phan \\
\\
Abstractive \\
\midrule
Best Previous & {\bf 41.69}$^\dagger$ & {\bf 19.47}$^\dagger$ & {\bf 39.08}$^\ddagger$ & 21.00$^\diamond$ \\
\midrule
\basespk & 40.88\phan & 17.80\phan & 38.54\phan & 20.38\phan \\ 
\basespk $\times 2$  & 40.76\phan & 17.88\phan & 38.46\phan & 19.88\phan \\ 
\pragr & 41.23\phan & 18.07\phan & 38.76\phan & 20.57\phan \\ 
\pragd & {\bf 41.39}\phan & {\bf 18.30}\phan & {\bf 38.78}\phan & {\bf 21.70}\phan \\ 
\bottomrule
\end{tabular}
\caption{\label{tbl:summarization}Test results for the non-anonymized CNN/Daily Mail summarization task. We compare to extractive baselines, and the best previous abstractive results of $^\dagger$\citet{celikyilmaz2018communicating},
$^\ddagger$\citet{paulus2018summarization} and $^\diamond$\citet{chen2018fast}.
We bold our highest performing model on each metric, as well as previous work if it outperforms all of our models.
}
\end{table}


The pragmatic methods obtain improvements of 0.2--0.5 in \rouge scores  and 0.2--1.8 \meteor over the base \basespk model, with the distractor-based approach \pragd outperforming the reconstructor-based approach \pragr. 
\pragd 
is strong across all 
metrics, obtaining results competitive to the best previous abstractive systems.



\begin{figure*}[t]
\centering
\begin{subfigure}[b]{0.45\textwidth}
\centering
\includegraphics[width=\textwidth]{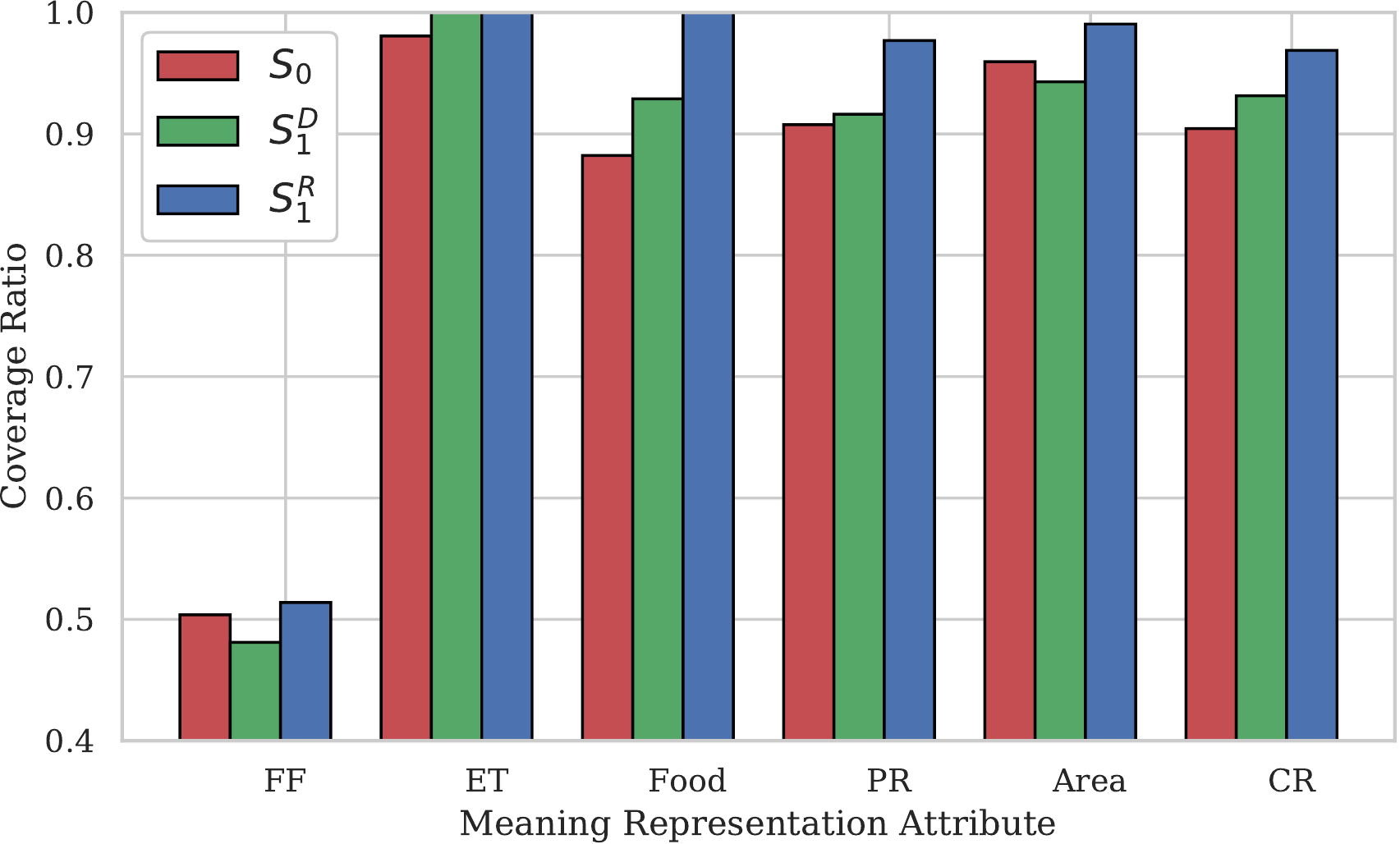}
\caption{
Coverage ratios by attribute type for the base model $S_0$ and pragmatic models $S_1^R$ and  $S_1^D$.
The pragmatic models typically improve coverage ratios across attribute types when compared to the base model.
\vspace{-.5em}
}
\end{subfigure}
\hspace{1em}
\begin{subfigure}[b]{0.49\textwidth}
\centering
\footnotesize
\scalebox{1.0}{
\begin{tabular}{p{1mm}l|cccccc}
\toprule
& & \multicolumn{6}{c}{Coverage Ratio for Attribute} \\
& & FF & ET & Food & PR & Area & CR \\
\midrule
& $S_0$ & 0.50 & 0.98 & 0.88 & 0.91 & 0.96 & 0.90 \\
\midrule
& \pragd-FF & \underline{0.57}\cellcolor{snsgreen!50} & 1.00\cellcolor{snsgreen!44} & 0.92\cellcolor{snsgreen!17} & 0.90\cellcolor{snsred!8} & 0.95\cellcolor{snsred!9} & 0.95\cellcolor{snsgreen!47} \\
& \pragd-ET & 0.47\cellcolor{snsred!24} & \underline{1.00}\cellcolor{snsgreen!50} & 0.96\cellcolor{snsgreen!32} & 0.92\cellcolor{snsgreen!12} & 0.96\cellcolor{snsred!2} & 0.95\cellcolor{snsgreen!50} \\
& \pragd-Food & 0.45\cellcolor{snsred!38} & 1.00\cellcolor{snsgreen!38} & \underline{1.00}\cellcolor{snsgreen!50} & 0.93\cellcolor{snsgreen!16} & 0.95\cellcolor{snsred!6} & 0.94\cellcolor{snsgreen!34} \\
& \pragd-PR & 0.51\cellcolor{snsgreen!3} & 1.00\cellcolor{snsgreen!50} & 0.90\cellcolor{snsgreen!9} & \underline{0.98}\cellcolor{snsgreen!50} & 0.93\cellcolor{snsred!31} & 0.92\cellcolor{snsgreen!18} \\
& \pragd-Area & 0.47\cellcolor{snsred!24} & 1.00\cellcolor{snsgreen!50} & 0.93\cellcolor{snsgreen!18} & 0.91\cellcolor{snsgreen!2} & \underline{0.98}\cellcolor{snsgreen!18} & 0.93\cellcolor{snsgreen!31} \\
 \rot{\rlap{Distractor Attr.}} & \pragd-CR & 0.45\cellcolor{snsred!37} & 1.00\cellcolor{snsgreen!44} & 0.91\cellcolor{snsgreen!12} & 0.90\cellcolor{snsred!4} & 0.91\cellcolor{snsred!50} & \underline{0.95}\cellcolor{snsgreen!45} \\
\bottomrule
\end{tabular}
}
\caption{
Coverage ratios by attribute type (columns) for the base model $S_0$, and for the pragmatic system $S_1^D$ when constructing the distractor by masking the specified attribute (rows). Cell colors are the degree the coverage ratio increases (green) or decreases (red) relative to $S_0$. 
\vspace{-.5em}
}
\end{subfigure}
\caption{Coverage ratios for the E2E task by attribute type, estimating how frequently the values for each attribute from the input meaning representations are mentioned in the output text. 
}
\vspace{-1em}
\label{fig:breakdown}
\end{figure*}

\section{Analysis}
The base speaker $S_0$ model is often underinformative, \eg for the E2E task failing to mention certain attributes of a MR, even though almost all the training examples incorporate all of them.
To better understand the performance improvements from the pragmatic models for E2E, we 
compute a \emph{coverage ratio} as a proxy measure of how well content in the input is preserved in the generated outputs.
The coverage ratio for each attribute is the fraction of times there is an exact match between the text in the generated output and the attribute's value in the source MR (for instances where the attribute is specified).\footnote{Note that this measure roughly provides a lower bound on the model's actual informativeness for each attribute, since the measure does not assign credit for paraphrases. 
}

Figure~\ref{fig:breakdown}(a) shows coverage ratio by attribute category for 
all models. 
The \pragr model increases the coverage ratio when compared to $S_0$ across all attributes, showing that using the reconstruction model score to select outputs does lead to an increase in mentions for each attribute. 
Coverage ratios increase for \pragd as well in four out of six categories, but the increase is typically less than that produced by \pragr. 

While \pragd optimizes less explicitly for attribute mentions than \pragr, it still provides a potential method to control generated outputs by choosing alternate distractors. 
Figure~\ref{fig:breakdown}(b) shows coverage ratios for \pragd when masking only a single attribute in the distractor.
The highest coverage ratio for each attribute is usually obtained when masking that attribute in the distractor MR (entries on the main diagonal, underlined), in particular for  \textsc{FamilyFriendly} (\textsc{FF}), \textsc{Food}, \textsc{PriceRange} (\textsc{PR}), and \textsc{Area}. However, masking a single attribute sometimes results in decreasing the coverage ratio, and we also observe substantial increases from masking other attributes: 
\eg masking either \textsc{FamilyFriendly} or \textsc{CustomerRating} (\textsc{CR}) produces an equal increase in coverage ratio for the \textsc{CustomerRating} attribute. This may reflect underlying correlations in the training data, as these two attributes have a small number of possible values (3 and 7, respectively). 

\section{Conclusion}

Our results show that
\basespk models from previous work, while strong,
still imperfectly capture the behavior that people exhibit when generating text; 
and an explicit pragmatic modeling procedure can improve results.
%
Both pragmatic methods evaluated in this paper encourage prediction of outputs that can be used to identify their inputs, either by reconstructing inputs in their entirety or distinguishing true inputs from distractors, so it is perhaps unsurprising that both methods produce similar improvements in performance. 
Future work might allow finer-grained modeling of the tradeoff between \emph{under}- and \emph{over}-informativity within the sequence generation pipeline (\eg with a learned communication cost model) or explore applications of pragmatics for content selection earlier in the generation pipeline.





\section*{Acknowledgments}
Thanks to Reuben Cohn-Gordon for many helpful discussions and suggestions. This work was supported
by DARPA through the XAI program. DF is supported by a Tencent AI Lab Fellowship.

\bibliography{refs}
\bibliographystyle{acl_natbib}

\appendix


\clearpage

\section{Supplemental Material}
\label{sec:supplemental}
\subsection{Reconstructor Model Details}
\label{sec:appendix_reconstructor}
 For the reconstructor-based speaker in the E2E task, we first follow the same data preprocessing steps as \citet{puzikov2018e2e}, which includes a delexicalization module that deals with sparsely occurring MR attributes (\textsc{name}, \textsc{near}) by mapping such values to placeholder tokens.
 
MRs have only a few possible values for most attributes: six out of eight attributes have fewer than seven unique values, and the remaining two attributes (\textsc{name}, \textsc{near}) are handled by our \basespk and \pragd using delexicalized placeholders, following \citet{puzikov2018e2e}.
In this way, the reconstructor only needs to predict the presence of these two attributes with a boolean variable, and other attributes with the corresponding categorical variable. We use a one layer bi-directional GRU \cite{cho2014gru} for the shared sentence encoder. We concatenate the latent vectors from both directions to construct a bi-directional encoded vector $h_i$ for every single word vector $d_i$ as:
\begin{subequations}
\begin{equation*}
\mathop{h_i}\limits ^{\rightarrow}=\mathop{GRU}\limits ^{\longrightarrow}(d_i, h_{i-1}), 
\mathop{h_i}\limits ^{\leftarrow}=\mathop{GRU}\limits ^{\longleftarrow}(d_i, h_{i+1})
\end{equation*}
\begin{equation*}
h_i = [\mathop{h_i}\limits ^{\rightarrow}, \mathop{h_i}\limits ^{\leftarrow}], i \in [1,L]
\end{equation*}	
\end{subequations}
Since not all words contribute equally to predicting each MR attribute, we thus use an attention mechanism~\cite{bahdanau2014neural} to determine the importance of every single word. The aggregated sentence vector for task $k$ is calculated by
\begin{equation*}
a_i^{(k)} = \frac{exp(W_a^{(k)}h_i)}{\sum_{j=1}^Lexp(W_a^{(k)}h_j)}, \\
v^{(k)}=\sum_{i=1}^La_i^{(k)}h_i,
\end{equation*}
The task-specific sentence representation is then used as input to $k$ layers with softmax outputs, returning a probability vector $Y^{(k)}$ for each of the $k$ MR attributes. 

\subsection{Hyperparameters}
\label{sec:appendix_hyperparameters}
For structured generation, we use beam size 10, $\lambda = 0.4$, and  $\alpha = 0.2$, tuned to maximize the normalized average of all five metrics on the development set.

For abstractive summarization, we use beam size 20, $\lambda = 0.9$, and  $\alpha = 1.0$, tuned to maximize \rouge-L on the development set.




\end{document}